\documentclass[dvipsnames,format=sigconf,nonacm]{acmart}

\usepackage{booktabs} 
\usepackage{amsmath, amsfonts}
\usepackage[utf8]{inputenc}
\usepackage{calrsfs}
\usepackage{prettyref}
\usepackage{algorithm}
\usepackage{algorithmic}
\usepackage{float}
\usepackage{graphicx}
\usepackage{acronym}
\usepackage{textcomp}
\usepackage{xcolor}
\usepackage{multirow}
\usepackage[caption=false]{subfig}
\usepackage[frozencache]{minted}
\usepackage{parcolumns}
\usepackage{microtype}
\usepackage{todonotes}
\usepackage{xspace}

\def\EvoAl{\textsc{EvoAl}\xspace}

\definecolor{bluegray}{RGB}{22,76,153}         
\definecolor{lightbluegray}{RGB}{215,225,255} 
\definecolor{greengray}{RGB}{22,153,76} 

\acrodef{SVR}[SVR]{Support Vector Regression}
\acrodef{EA}[EA]{Evolutionary Algorithm}
\acrodef{DSL}[DSL]{domain-specific language}

\newrefformat{sec}{Section~\ref{#1}}
\newrefformat{alg}{Algorithm~\ref{#1}}
\newrefformat{fig}{Figure~\ref{#1}}
\newrefformat{tab}{Table~\ref{#1}}
\newrefformat{lst}{Listing~\ref{#1}}
\newrefformat{eq}{Equation~\ref{#1}}
\newrefformat{rq}{Research Question~\ref{#1}}

\copyrightyear{2024}
\acmYear{2024}
\acmConference[GECCO '24]{Genetic and Evolutionary Computation Conference Companion}{July 15--19, 2024}{Lisbon, Portugal}

\begin{document}
  \title{$\EvoAl^{2048}$}

  \author{Bernhard J. Berger}
  \affiliation{%
    \institution{University of Rostock, Software Engineering Chair}  
    \city{Rostock}
    \country{Germany}}
  \additionalaffiliation{%
  \institution{Hamburg University of Technology, Institute of Embedded Systems}  
  \city{Hamburg}
  \country{Germany}}
  \email{bernhard.berger@uni-rostock.de}

  \author{Christina Plump}
  \affiliation{%
    \institution{DFKI --- Cyber-Physical Systems}
    \streetaddress{Bibliotheksstraße 5}
    \city{Bremen}
    \country{Germany}}
  \email{Christina.Plump@dfki.de}

  \author{Rolf Drechsler}
  \affiliation{%
    \institution{University of Bremen, Departments of Mathematics and Computer Science}
    \streetaddress{Bibliotheksstraße 5}
    \city{Bremen}
    \country{Germany}}
  \additionalaffiliation{%
    \institution{DFKI --- Cyber-Physical Systems}
    \streetaddress{Bibliotheksstraße 5}
    \city{Bremen}
    \country{Germany}}
  \email{drechsler@uni-bremen.de}

  \maketitle

\section{introduction}
  Explainability and interpretability of solutions generated by AI products are getting more and more important as AI solutions enter safety-critical products.
  As, in the long term, such explanations are the key to gaining users' acceptance of AI-based systems' decisions~\cite{8494900}.

  We report on the application of a model-driven optimisation to search for an interpretable and explainable policy that solves the game \emph{2048}.
  This paper describes a solution to the \emph{Interpretable Control Competition}~\cite{nadizar:2024}.
  We focus on solving the discrete \emph{2048} game challenge using the open-source software \EvoAl~\cite{10253985,evoal:homepage:2024} and aimed to develop an approach for creating interpretable policies that are easy to adapt to new ideas. We use a model-driven optimisation approach~\cite{John2023} to describe the policy space and use an evolutionary approach to generate possible solutions.
  Our approach is capable of creating policies that win the game, are convertible to valid Python code, and are useful in explaining the move decisions.

  \section{Approach}\label{sec:tooling}
  The proposed solution builds on \EvoAl---a Java-based data-science research
  tool---which allows users to express optimisation problems using domain-specific
  languages (DSLs) and offers a rich extension API for problem-specific extensions.
  \EvoAl offers different optimisation algorithms, such as evolutionary algorithms,
  genetic programming, and model-driven optimisation.

  Normally, \EvoAl uses two DSLs to configure an optimisation problem. Using the
  \emph{data description language}, a user can specify the problem-specific data.
  The mapping to an optimisation algorithm and the algorithm configuration is done
  by using the \emph{optimisation language}. As we aim to use a model-driven
  approach, we use a third DSL of \EvoAl---the \emph{definition language}---to
  describe the abstract syntax~\cite{Combemale:684289} of the policy. The generated
  model is then turned into Python code by using model-to-text concepts. 

  \begin{figure}[ht]
    \centering
    \includegraphics[width=.8\columnwidth]{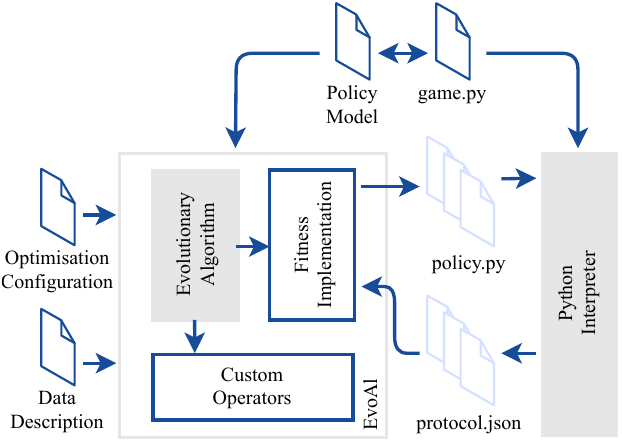}
    \caption{Depiction of our generation process}
    \label{fig:approach}
    \Description{Policy model overview} 
  \end{figure}

  Figure~\ref{fig:approach} shows a sketch of our approach. We provide three DSL
  files to \EvoAl, the data description, optimisation configuration, and the
  policy model. The policy model is linked to the Python module (\emph{game.py}),
  which implements game state queries the policy can use. Furthermore, we implement
  \EvoAl extensions for the fitness calculation and additional problem-specific
  operators for the evolutionary algorithm, which is part of \EvoAl. The fitness
  calculation generates a Python module (\emph{policy.py}) for each individual
  that implements its policy. Then, the Python interpreter runs a configurable
  number of games and writes the results to log files (\emph{protocol.json}),
  which are then read by the fitness calculation and passed to the EA.

\begin{figure}[ht]
    \centering
    \includegraphics[width=.8\columnwidth]{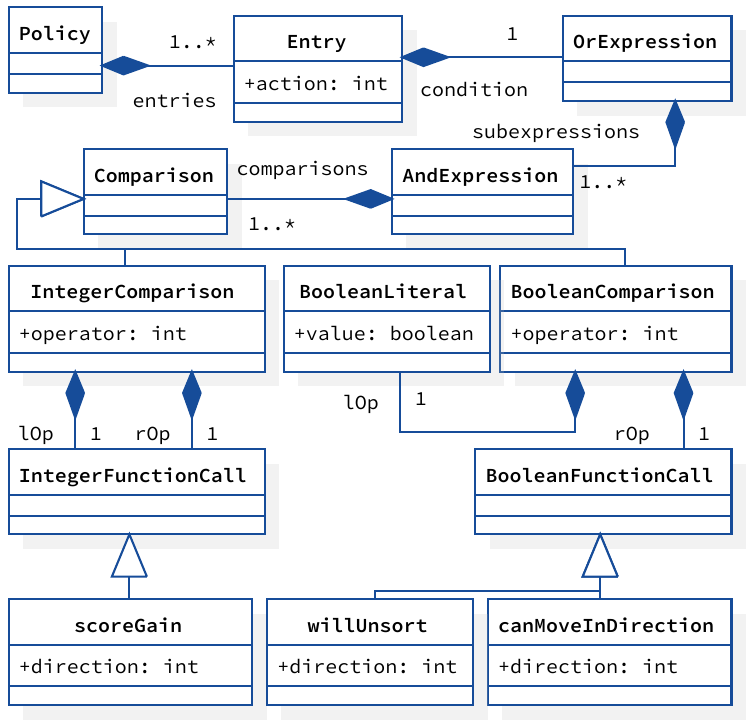}
    \caption{Excerpt of our policy model}
    \label{fig:policy-model}
    \Description{The policy model} 
  \end{figure}

  Our policy model builds upon the idea that a game is in a state, which
  influences the action taken. The state can be queried by simple functions,
  such as \emph{Is a certain move valid?}, or \emph{What is the gain of a specific
  move?}. A policy then combines these functions into boolean expressions, which are
  checked to determine if a specific action should be executed.
  Figure~\ref{fig:policy-model} shows an excerpt of the abstract syntax of our
  policy model, which we describe for \EvoAl using the mentioned \emph{description
  languge}. By using this approach, it is possible to add new query functions by
  a) implementing the Python code and b) extending the description file. It is
  not necessary to adapt \EvoAl or our Java-based extension to do so.

  \noindent\textbf{Initial Population} The initial population contains random
  policies containing a single element in array-like references, c.f.
  Figure~\ref{fig:policy-model}. Thus, the policies start with simple conditions
  that need to be mutated into more and more complex ones.  

  \noindent\textbf{Operators} We mainly employ general (not specialised for the
  problem) operators. We use a mutator that changes values (numbers, and boolean
  values), a mutator that changes array sizes (by removing or adding elements),
  a mutator that changes the order of arrays, and a custom mutator that rotates a
  given policy. We limit the number of mutators applied to an offspring to one,
  to reduce the changes in a policy. Additionally, we use a standard recombination 
  operator, that swaps subtrees of two provided policies.
  
  \noindent\textbf{Fitness} Our fitness function executes a policy a configurable
  number of times and calculates different statistical information, such as the
  minimum, maximum, and average of these runs. For one run, we store the maximum
  tile value and the overall score. This allows us to focus on searching
  for robust algorithms. We use a priority-ordered pareto-comparison as a fitness
  comparator.

  \noindent\textbf{Available Functions}
  We aimed at using query functions that are simple and do not implement complex
  board situations. In total, we provide ten different query functions, such as
  \emph{canMoveInDirection} allows the policy to query if a
  single move into a direction is possible, whereas \emph{canMoveInDirections}
  checks for two subsequent moves. \emph{scoreGain} will calculate the score
  improvement gained by a single move, whereas \emph{scoreGains} checks two
  subsequent moves. The complete list of functions can be found in the policy
  model, which is specified in the file \emph{model.dl}.


  \noindent\textbf{Pipeline} The pipeline configuration is stored in the
  aforementioned DSL files, which can be found in the folder
  \texttt{evoal-configuration} of the supplemental repository~\cite{berger:gh:2024}.
  The files can be read with a text editor. \EvoAl's Eclipse-based DSL
  editors provide additional syntax highlighting and cross-referencing.
  The pipeline has been tested on Linux and MacOS using a \emph{Java 17 JRE}.
  To run the pipeline, it is necessary to checkout the repository, set up the
  Python environment, download an \EvoAl release~\cite{berger:evoal:2024} and
  extract it to the folder \texttt{evoal-release} in the repository and
  execute the script file \texttt{01-run-search.sh}. 

  \section{Experimental Results}\label{sec:results}
  The allowed budget contains $200.000$ evaluations of the game. We configured
  the EA to use a population size of 100 individuals. For a fitness evaluation,
  we decided to simulate six games with the same policy, resulting in
  $\frac{200000}{100\cdot 6} = 333$ generations that can be executed.

\begin{figure}[h]
    \centering
    \includegraphics[width=\columnwidth]{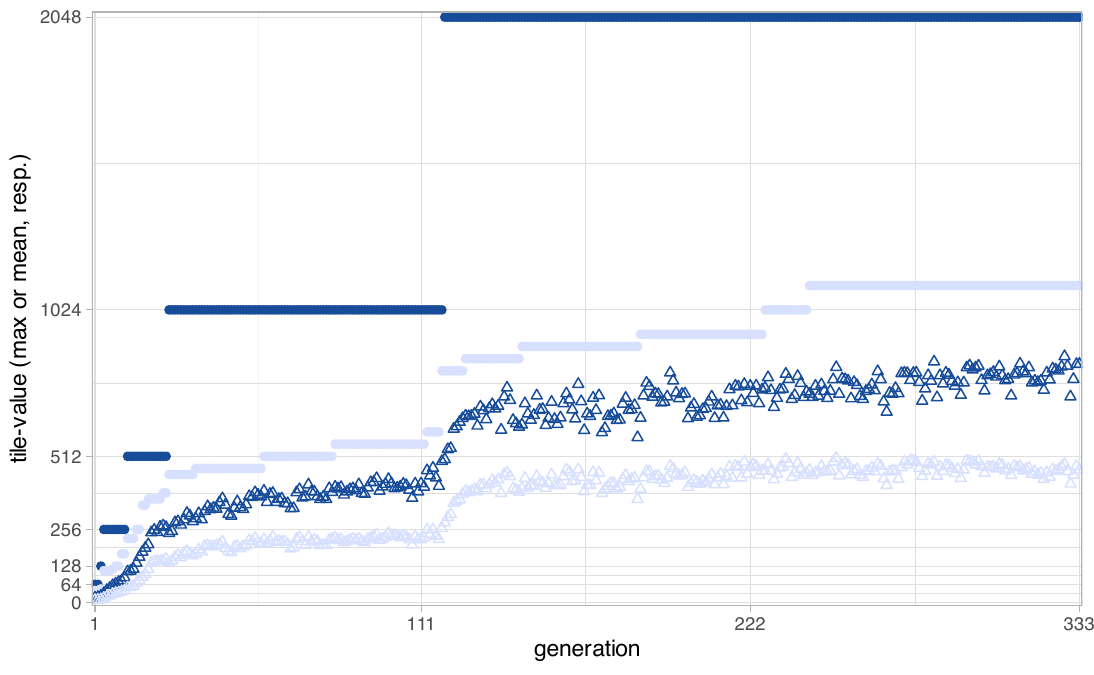}
    \caption{Overview of the fitness values over time}
    \label{fig:score-development}
    \Description{} 
  \end{figure}

  \noindent\textbf{Best Policy}
  Figure~\ref{fig:score-development} shows the development of the highest tile
  reached during the evolution process. As a policy run simulates six games, the
  data points in dark blue show the best highest tile value of a policy run and
  the light blue data points show the average value of the highest tile of a
  policy run. The filled data points represent the best individual of a generation,
  while the non-filled data points represent the average result of a generation. 
  The depiction shows that whenever the process succeeds in generating a new best
  highest tile, the average highest tile first improves (the policies are getting
  more stable) before the best highest tile can reach the next level. 

  The result of the optimisation run is a policy that reached a $max(highest{-}tile)$ of $2.048$,
  an $average(highest{-}tile)$ of $1.276$, and an $average$ $(total{-}score)$ of 
  $14.093$. The policy is, during the optimisation, an instantiation graph of
  the model shown in Figure~\ref{fig:policy-model}, which can be converted into
  a graphical representation or a textual representation. Nevertheless,
  Listing~\ref{lst:policy} shows the best policy after converting it into a
  Python program as this is the executable policy.

  \begin{listing}[hb]
\begin{minted}{python3}
import numpy as np
import evoal.game as game


def control(observation: np.ndarray) -> int:
  """Generated policy function for the game 2048"""

  if game.scoreGains(observation, direction1 = 2, direction2 = 3) > game.scoreGains(observation, direction1 = 1, direction2 = 0) or game.scoreGains(observation, direction1 = 2, direction2 = 2) > game.scoreGains(observation, direction1 = 3, direction2 = 2):
    return game.toAction(2)
  if game.scoreGains(observation, direction1 = 1, direction2 = 0) < game.scoreGains(observation, direction1 = 3, direction2 = 2) or not game.canMoveInDirection(observation, direction = 1):
    return game.toAction(3)
  if game.scoreGains(observation, direction1 = 2, direction2 = 3) > game.scoreGains(observation, direction1 = 1, direction2 = 1) or game.scoreGains(observation, direction1 = 2, direction2 = 1) > game.scoreGains(observation, direction1 = 0, direction2 = 1):
    return game.toAction(2)

  return game.toAction(1)
\end{minted}
    \caption{Best policy after 333 generations}
    \label{lst:policy}
  \end{listing}

  The shown policy focuses on increasing the score gain and chooses to go in the direction that promises higher score gain.
  The remaining queries, such as $willBeSorted$, are part of some policies but did not make it into the best policy.
  At the same time, the policy only uses three out of four directions, which might leave room for further improvement, but we assume that the situation where the board would have to be moved into the fourth direction occurs very seldom.

  Having a given board situation, the policy allows one to explain precisely why a certain move was made.
  On the one hand, the state queries are easy to understand and, on the other hand, they can be calculated for a given board to show the decision process.
  While being explainable, our approach is flexible and can easily be extended with additional queries without having to change the optimisation process.
  
  \bibliographystyle{ACM-Reference-Format}
  \bibliography{literature/bibliography.bib}
\end{document}